\documentclass[sn-mathphys]{sn-jnl}

\usepackage{amsmath}
\usepackage{mathtools}
\usepackage{multirow}
\usepackage{makecell}
\usepackage{float}

\jyear{2026}%
\theoremstyle{thmstyleone}%

\theoremstyle{thmstyletwo}%

\theoremstyle{thmstylethree}%

\makeatletter
\def\segsrunningtitle{}
\def\ps@titlepage{%
  \def\@oddhead{\hbox to \hsize{\headerfont\segsrunningtitle\hfill\thepage}}%
  \let\@evenhead\@oddhead
  \let\@oddfoot\@empty
  \def\@evenfoot{}%
}
\def\ps@headings{%
  \def\@oddhead{\hbox to \hsize{\headerfont\segsrunningtitle\hfill\thepage}}%
  \def\@evenhead{\hbox to \hsize{\headerfont\thepage\hfill\segsrunningtitle}}%
  \let\@oddfoot\@empty
  \let\@evenfoot\@empty
  \let\@mkboth\markboth%
}
\makeatother

\begin{document}

\title[Structural Energy Guidance for View-Consistent Text-to-3D Generation]
{Structural Energy Guidance for View-Consistent Text-to-3D Generation}

\author[1]{\fnm{Qing} \sur{Zhang}}

\author[1,2]{\fnm{Jinguang} \sur{Tong}}

\author*[1]{\fnm{Jing} \sur{Zhang}}
\email{jing.zhang@anu.edu.au}

\author*[3]{\fnm{Jie} \sur{Hong}}
\email{jiehong@hku.hk}

\author*[1,2]{\fnm{Xuesong} \sur{Li}}
\email{xuesong.li@csiro.au}

\affil[1]{\orgname{Australian National University},
\orgaddress{\city{Canberra}, \postcode{2601}, \state{ACT}, \country{Australia}}}

\affil[2]{\orgname{CSIRO},
\orgaddress{\city{Canberra}, \postcode{2601}, \state{ACT}, \country{Australia}}}

\affil[3]{\orgname{The University of Hong Kong},
\orgaddress{\city{Hong Kong SAR}}}

\abstract{Text-to-3D generation based on diffusion models often suffers from the Janus problem, leading to inconsistent geometry across viewpoints. This work identifies viewpoint bias in 2D diffusion priors as the main cause and proposes Structural Energy-Guided Sampling (SEGS), a training-free and plug-and-play framework to improve multi-view consistency. SEGS constructs a structural energy in the PCA subspace of U-Net features and injects its gradient into the denoising process. It can be easily integrated into SDS/VSD pipelines without retraining. Experiments show that SEGS reduces the Janus Rate by about 10\% on average and improves View-CS scores across multiple baselines, including DreamFusion, Magic3D, and LucidDreamer. This method effectively alleviates viewpoint artifacts while preserving appearance fidelity, providing a flexible solution for high-quality text-to-3D content generation.}

\keywords{Text-to-3D Generation, Diffusion Models, Score Distillation Sampling, View Consistency, 3D Generation}

\maketitle

\section{Introduction}
\label{sec:intro}
High-quality 3D assets are foundational to entertainment, product design, AR/VR, and simulation. Despite the rapid progress of text-to-image generation driven by large-scale web datasets~\cite{schuhmann2022laion}, the scarcity and cost of broad-coverage 3D supervision~\cite{deitke2024objaverse} still constrain text-to-3D pipelines. The strong transferability of foundation models in vision domains~\cite{wu2025eyecare} further motivates the reuse of pretrained priors for downstream tasks. This raises a central question: how can we transfer the rich priors of powerful 2D diffusion models into reliable, multi-view consistent 3D representations without relying on large curated 3D corpora? A seminal step in this direction is DreamFusion~\cite{poole2022dreamfusion}, which optimizes a Neural Radiance Field (NeRF) from a frozen text-to-image diffusion prior via Score Di
stillation Sampling (SDS). The SDS paradigm has since inspired numerous systems~\cite{chen2023fantasia3d, lin2023magic3d, han2024latent, liang2024luciddreamer, zhang2025improving, han2025enhancing, wang2024prolificdreamer, cao2024dreamavatar}, making high-fidelity text-conditioned synthesis feasible without paired 3D training data.
\begin{figure}[h]
    \centering
       \includegraphics[width=0.86\linewidth]{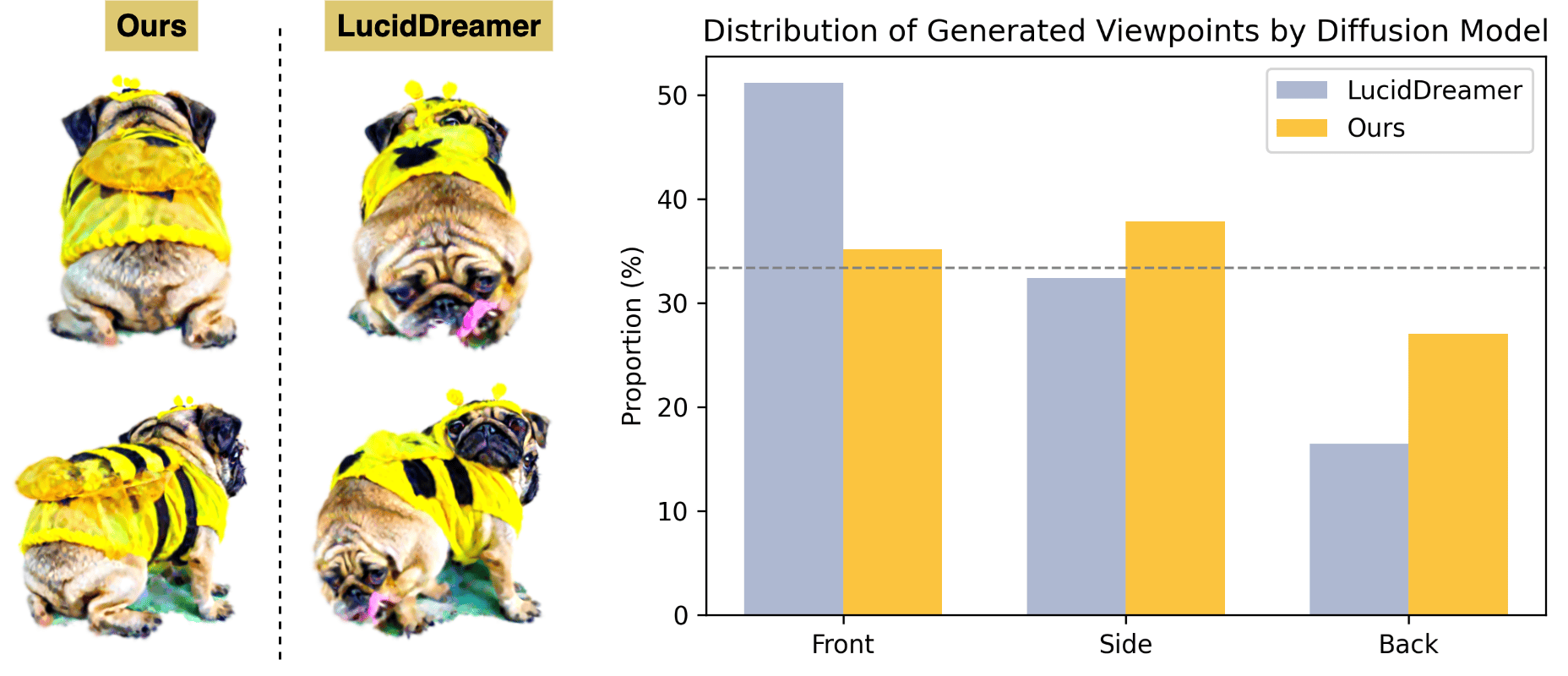}
    \caption{
        Comparison of view distributions generated by LucidDreamer~\cite{liang2024luciddreamer} and SEGS, using the prompt ``a DSLR photo of a pug wearing a bee costume.'' SEGS yields a more uniform view sampling distribution, whereas LucidDreamer exhibits a typical long-tail pattern. As a result, Janus artifacts are reduced in the SEGS output.}
    \label{fig:example_motivation}
\end{figure}

Despite the practical success of SDS-style pipelines, they frequently suffer from the Janus problem: shapes that appear plausible from a canonical front view deteriorate into duplicated or distorted geometry from other angles (Figure~\ref{fig:example_motivation}). One line of remedies replaces the frozen 2D prior with a multi-view–adapted version, by retraining or fine-tuning on curated view supervision, as in Zero-1-to-3 and MVDream~\cite{liu2023zero, shi2023mvdream}. Although this improves pose coverage, updating the score network reshapes the appearance manifold: high-frequency textures and material cues are weakened, styles drift toward the adapted domain, and cross-category generalization declines~\cite{huang2024dreamcontrol}. Moreover, the supervision, curation, and compute overhead reduce the plug-and-play appeal of SDS/VSD. Other attempts instead steer generation with external proxies such as edges, depth, sketches, or layout, using control branches or multi-view conditioning~\cite{huang2024dreamcontrol}.

\begin{figure}[t]
    \centering
    \includegraphics[width=\linewidth]{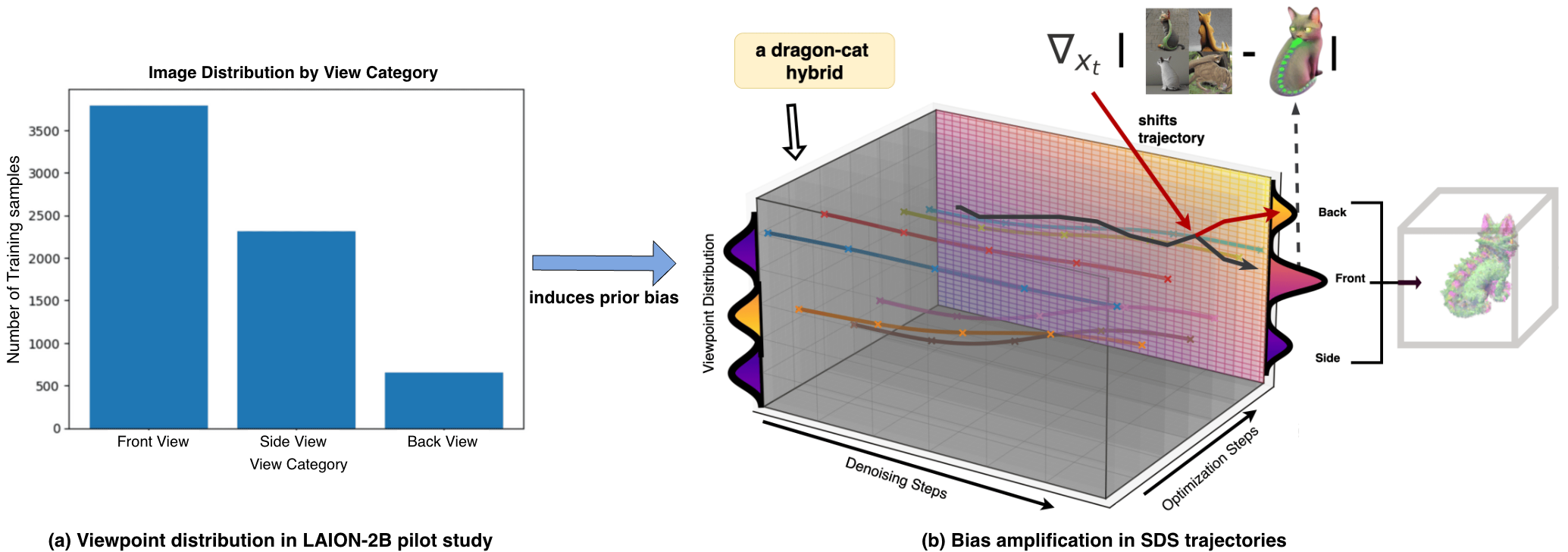}
    \caption{\textbf{Viewpoint bias and its amplification in SDS.}
    (a) A pilot analysis on LAION-2B shows a skewed viewpoint distribution, where frontal views are over-represented, side and back views are under-represented.
    (b) During SDS optimization, this data bias is inherited by the diffusion prior and pulls denoising trajectories toward the frontal mode, amplifying view inconsistency in text-to-3D generation.}
    \label{fig:motivation}
\end{figure}

Text-to-image diffusion priors trained on Internet photos exhibit a viewpoint bias toward frontal views. SDS optimizes the 3D representation so that renders from random viewpoints are consistently scored as high-likelihood images by the frozen prior. This process unintentionally reinforces the prior’s frontal preference across all poses, ultimately leading to Janus artifacts. 
Figure~\ref{fig:motivation} summarizes this mechanism: a skewed training-view distribution induces a biased diffusion prior, which is then amplified by SDS trajectories during 3D optimization.
Prior work has shown that intermediate diffusion features, including self-attention representations, encode spatially aligned object structure, which is useful for downstream control and editing~\cite{tumanyan2023plug,mo2024freecontrol}.
Our key insight is to expose this structural information as an energy-like target and use it to steer the sampling trajectory toward the intended structural configuration of the target view, counteracting the bias while preserving fine textures and material details.

To address this issue, we propose \textbf{SEGS} (Structural Energy-Guided Sampling), a training-free approach that improves multi-view consistency during generation while preserving fine appearance. SEGS offers two complementary signals: \textit{Structural Energy Guidance}, which exposes the structure encoded in intermediate U\hbox{-}Net activations by constructing a viewpoint-aware PCA subspace and defining a simple structural energy whose gradient is injected into the denoising update to steer the sampling trajectory toward the target view while keeping the diffusion weights frozen (Figure~\ref{fig:pipeline}); and a lightweight \textit{Text Consistency Guard}, which optionally prunes supervision that is inconsistent with the requested viewpoint to stabilize edge cases. Operating purely at sampling time, SEGS integrates seamlessly with SDS/VSD and controls the trajectory rather than the parameters.
In summary, our contributions are as follows:
\begin{itemize}
\item We introduce \textbf{SEGS}, a sampling-time formulation that casts view consistency as minimizing a structural energy in a PCA subspace of intermediate diffusion features, steering geometry toward the target view without changing model weights.
\item The approach is \textbf{training-free} and \textbf{plug-and-play}, integrates with SDS/VSD, and requires \textbf{no additional predictors}, preserving high-frequency textures and material fidelity.
\item Across strong text-to-3D baselines, SEGS reduces the Janus problem and improves multi-view alignment under the same compute budget.
\end{itemize}

\section{Related Work}
\label{sec:related_work}
\subsection{Paradigms of Text-to-3D Generation}

Existing text-to-3D frameworks broadly fall into three paradigms: end-to-end generation, SDS-based 2D-to-3D distillation, and reconstruction-based generation.

\noindent \textbf{End-to-End Generation.} 
Early methods trained a single network to directly map text embeddings into 3D representations, such as Point-E~\cite{nichol2022point} and Shap-E~\cite{jun2023shap}. While conceptually simple and fast at inference, these approaches remain limited in fidelity and category coverage due to the scarcity of large-scale 3D data~\cite{liu2024comprehensive}, motivating the shift toward leveraging 2D priors.

\noindent \textbf{SDS-based 2D-to-3D Distillation.} 
DreamFusion~\cite{poole2022dreamfusion} introduced Score Distillation Sampling (SDS), which quickly became the dominant paradigm by using frozen 2D diffusion priors to optimize NeRFs~\cite{mildenhall2021nerf, xiao2025neural}. 
Subsequent works improved quality (e.g., ProlificDreamer, Fantasia3D, SwiftCraft3D) or speed, such as GaussianDreamer and DreamGaussian, which exploit Gaussian Splatting for fast rendering~\cite{wang2024prolificdreamer, chen2023fantasia3d,tang2023dreamgaussian, kerbl20233d, tong2025gs, li2025dgns, dong2026swiftcraft3d, yi2024gaussiandreamer}. However, because they inherit viewpoint biases from the underlying 2D models, these pipelines still suffer from Janus artifacts.
Related text-guided 3D manipulation has also been explored through diffusion-based mesh deformation and CLIP-guided face stylization~\cite{xu2024fusiondeformer,gao2025disentangled}.

\noindent \textbf{Reconstruction-based Generation.} 
Recent methods such as Instant3D~\cite{li2023instant3d} and LGM~\cite{tang2024lgm} regress 3D assets from sparse multi-view images synthesized by fine-tuned diffusion models. These approaches are efficient and high-resolution, yet still rely on biased 2D priors, leading to similar multi-view inconsistencies. 

\subsection{View Consistency in Text-to-3D}
Geometric consistency has also been explored in broader 3D vision tasks such as non-rigid point cloud registration~\cite{wang2025non}. In text-to-3D generation, efforts to address view inconsistency fall into three directions: retraining diffusion priors, injecting external supervision, and sampling-time guidance.

\noindent \textbf{Retraining or Fine-tuning.}
Methods such as Zero-1-to-3~\cite{liu2023zero} and MVDream~\cite{shi2023mvdream} enhance multi-view awareness by fine-tuning diffusion models on curated multi-view datasets with 3D-aware attention. While effective, these approaches rely on domain-specific 3D corpora—often stylized or synthetic—which leads to weakened textures, reduced detail fidelity, and poorer cross-category generalization~\cite{huang2024dreamcontrol}. They also require additional compute, undermining the plug-and-play appeal of SDS pipelines.

\noindent \textbf{External Supervision.}
Another line introduces proxy signals, such as edges, depth, or sketches, via ControlNet-style conditioning~\cite{huang2024dreamcontrol, chen2023control3d}. These methods stabilize geometry but demand extra predictors and training, and they do not provide explicit structural targets for viewpoint alignment.

\noindent \textbf{Sampling-time Guidance.}
A third direction keeps the backbone frozen and modifies the guidance signal at inference. For example, D-SDS~\cite{hong2023debiasing} debiases the score and prompt, and Perp-Neg~\cite{armandpour2023re} reparameterizes negative prompts to mitigate semantic conflicts. While lightweight and plug-and-play, such methods operate only in text/score space, without offering structural cues.

\section{Preliminaries}
\label{sec:pre}
\subsection{Score Distillation Sampling}
Let $x_0=g(\theta,v)$ denote the rendered image of a differentiable 3D representation with parameters $\theta$ under camera view $v$. Let $\epsilon_\phi(x_t,t,y)$ be the noise prediction of a pretrained text-to-image diffusion model, conditioned on the noisy image $x_t$, timestep $t$, and text prompt $y$. The Score Distillation Sampling (SDS) gradient~\cite{poole2022dreamfusion} is
\begin{equation}
\nabla_{\theta} \mathcal{L}_{\text{SDS}}(\theta) \approx 
\mathbb{E}_{t,\epsilon,v} \!\left[ \,\omega(t)\,\big(\epsilon_{\phi}(x_t,t,y) - \epsilon_t\big)\,\frac{\partial x_0}{\partial \theta} \right],
\label{eq:sds_gradient}
\end{equation}
where $\epsilon_t$ is Gaussian noise and $\omega(t)$ is a timestep weighting.
From DDPM~\cite{ho2020denoising,song2020denoising}, the clean-image estimate is
\begin{equation}
 \hat{x}_0^t\ = \frac{x_t - \sqrt{1 - \bar{\alpha}_t}\,\epsilon_{\phi}(x_t,t,y)}{\sqrt{\bar{\alpha}_t}},
\end{equation}
with $\bar{\alpha}_t=\prod_{i=1}^{t}\alpha_i$ and $\gamma(t)=\sqrt{(1-\bar{\alpha}_t)/\bar{\alpha}_t}$. An equivalent form of Eq.~\eqref{eq:sds_gradient} can be written as Eq.~\eqref{eq:sds_pseudotarget}:
\begin{equation}
\nabla_{\theta} \mathcal{L}_{\text{SDS}}(\theta) =
\mathbb{E}_{t,\epsilon,v} \!\left[ \frac{\omega(t)}{\gamma(t)}\,\big(x_0-\hat{x}_0^t\big)\,\frac{\partial x_0}{\partial \theta} \right].
\label{eq:sds_pseudotarget}
\end{equation}

This formulation provides a standard interface for modifying the denoising prediction used by SDS, which we will exploit to introduce a structure-aware trajectory correction.

\subsection{Energy-Guided Denoising}
Following the general principle of energy-based guidance in diffusion sampling~\cite{dhariwal2021diffusion,epstein2023diffusion}, we adopt an energy-based view of controllable sampling. At each denoising step, an additional, task-specific directional term derived from a differentiable \emph{energy} is injected into the diffusion model’s noise prediction to steer the sampling trajectory while keeping the backbone frozen:
\begin{equation}
\hat{\epsilon}_\phi(x_t,t,y)\;=\;\epsilon_\phi(x_t,t,y)\;+\;\lambda_v(t)\,\nabla_{x_t}E(x_t,t),
\label{eq:eg_overview}
\end{equation}
where $E(x_t,t)$ is designed such that lower values correspond to more desirable samples, and $\lambda_v(t)$ controls the guidance schedule.
Note that since DDPM/DDIM updates subtract $\hat{\epsilon}_\phi$ from the state, the $+\,\nabla_{x_t}E$ term in Eq.~\eqref{eq:eg_overview} induces a descent step on $E$ at the state level.
This yields a training-free, plug-and-play mechanism to reshape the denoising path. In Section~\ref{sec:method}, we instantiate $E$ as a \emph{PCA-structural energy} defined in a viewpoint-aware structural subspace and detail its coupling to the SDS objective.

\section{Viewpoint Bias in Diffusion-based Text-to-3D}
\label{sec:viewpoint-bias}
This section analyzes how viewpoint bias arises in frozen 2D diffusion priors and how it is amplified by SDS during text-to-3D optimization. 
We first relate the learned diffusion score to the training-data distribution and provide pilot evidence of viewpoint imbalance in LAION-2B. 
We then show that SDS converts this prior bias into imbalanced pseudo-target supervision, which explains why camera sampling alone cannot eliminate Janus artifacts and motivates a sampling-time structural correction.

\subsection{Viewpoint Bias in Diffusion Priors}
We first clarify how viewpoint imbalance in training images can be inherited by a frozen diffusion prior. 
Let \(p_{\text{data}}(x_0)\) denote the distribution of training images, and let \(p_t(x)\) denote the marginal density of noisy samples \(x_t\) obtained by diffusing \(x_0 \sim p_{\text{data}}\) to timestep \(t\). 
Therefore, structural statistics of the training data, including viewpoint frequencies, can persist in \(p_t(x)\).

From this perspective, the forward diffusion process can be expressed as~\cite{song2020score}:
\begin{equation}
dx = f(x, t)\,dt + g(t)\,dW_t .
\end{equation}

The corresponding Fokker--Planck equation governing the density \( p_t(x) \) is~\cite{oksendal2013stochastic, risken1996fokker}
\begin{equation}
\frac{\partial p_t(x)}{\partial t} 
= -\nabla_x \cdot \big( f(x, t) p_t(x) \big) 
+ \tfrac{1}{2} g(t)^2 \nabla_x^2 p_t(x).
\end{equation}

Using the identity:
\begin{equation}
\nabla_x \log p_t(x) = \frac{\nabla_x p_t(x)}{p_t(x)},
\label{eq:score_identity}
\end{equation}
the diffusion term can be rewritten as:
\[
\tfrac{1}{2} g(t)^2 \nabla_x^2 p_t(x) 
= \tfrac{1}{2} g(t)^2 \nabla_x \cdot \big( p_t(x) \nabla_x \log p_t(x) \big).
\]

Substituting this back yields the compact form: 
\begin{equation}
\frac{\partial p_t(x)}{\partial t} 
= -\nabla_x \cdot \Big[ \big( f(x,t) - \tfrac{1}{2} g(t)^2 \nabla_x \log p_t(x) \big) p_t(x) \Big].
\label{eq:forward_fp}
\end{equation}

The key term is the score \(\nabla_x \log p_t(x)\), which points along gradients of the noisy training-data density and is the quantity approximated by score-based diffusion models.
Consequently, the learned denoising direction is tied to high-density modes of the training distribution rather than to an explicitly viewpoint-balanced objective.

The reverse diffusion SDE is given by~\cite{song2020score}
\begin{equation}
dx = \big[f(x, t) - g(t)^2 \nabla_x \log p_t(x)\big]\,dt + g(t)\,d\bar W_t .
\end{equation}

Introducing \( s = T - t \) with $ds=-dt$, we obtain
\begin{equation}
dx = \big[-f(x, t) + g(t)^2 \nabla_x \log p_t(x)\big]\,ds + g(t)\,dW_s .
\end{equation}

Here, \(dW_s\) denotes a standard Wiener process in the new time variable \(s\).

The corresponding Fokker--Planck equation for the density \( q_s(x) \) is
\begin{equation}
\begin{aligned}
\frac{\partial q_s(x)}{\partial s} 
&= -\nabla_x \cdot \Big\{ q_s(x) \big[ -f(x,t) + g(t)^2 \nabla_x \log p_t(x) \big] \Big\} \\
&\quad + \tfrac{1}{2} g(t)^2 \nabla_x^2 q_s(x).
\end{aligned}
\end{equation}

Using the same score identity as Eq.~\eqref{eq:score_identity} for \(q_s(x)\), this becomes
\begin{equation}
\begin{aligned}
\frac{\partial q_s(x)}{\partial s} = 
\nabla_x \cdot \Big[ \big( f(x,t) 
- g(t)^2 \nabla_x  \log p_t(x) 
\\ + \tfrac{1}{2} g(t)^2 \nabla_x \log q_s(x) \big) q_s(x) \Big].
\end{aligned}
\end{equation}

Under ideal reverse-time matching, the reverse-time density \(q_s(x)\) remains close to the corresponding forward-time density \(p_t(x)\), so their scores can be approximated as:
\[
\nabla_x \log q_s(x) \approx \nabla_x \log p_t(x),
\]
which leads to
\begin{equation}
\frac{\partial q_s(x)}{\partial s} 
= -\nabla_x \cdot \Big[ \big( \tfrac{1}{2} g(t)^2 \nabla_x \log p_t(x) - f(x,t) \big) q_s(x) \Big].
\label{eq:reverse_fp}
\end{equation}

Under this matching approximation, Equations~\eqref{eq:forward_fp} and~\eqref{eq:reverse_fp} are consistent with reverse-time probability flow. 
Thus, the reverse diffusion process tends to retrace the probability flow of the forward process, reconstructing samples that approximate the original data distribution \( p_{\text{data}}(x) \). 
The derivation should not be read as a proof of viewpoint bias; it clarifies how biases in \(p_{\text{data}}\) can enter the learned score field and reappear during reverse denoising.

To check whether such skew exists in the training corpus, we conduct a small pilot study on the LAION-2B dataset~\cite{schuhmann2022laion}, which serves as the training source for Stable Diffusion~\cite{rombach2022high}. 
We randomly sample $10{,}000$ images and apply a CLIP-based viewpoint classifier to assign coarse labels (front/side/back), discarding low-confidence predictions. 
As shown in Figure~\ref{fig:motivation}(a), the resulting distribution is clearly skewed toward frontal views. 
Although preliminary, this observation supports the hypothesis that viewpoint imbalance in Internet-scale training data can be inherited by frozen text-to-image diffusion priors.

\subsection{Bias Amplification in SDS Optimization}
We next examine how inherited viewpoint bias affects text-to-3D optimization. 
Although SDS provides a convenient interface for supervision (Eq.~\eqref{eq:sds_pseudotarget}), the quality of its guidance depends entirely on the pseudo-targets produced by the frozen prior. 
When this prior favors frontal views, the pseudo-targets $\hat{x}_0^t$ generated during optimization are not uniformly distributed across poses: frontal predictions can dominate even when camera poses are sampled uniformly.

To inspect this effect, we log $\hat{x}_0^t$ throughout SDS optimization and classify their viewpoints (front/side/back) with a CLIP-based classifier~\cite{radford2021learning}. 
The resulting statistics reveal a clear imbalance: frontal views are over-represented, while side and back views are under-sampled. 
A representative view-distribution comparison is shown in Figure~\ref{fig:example_motivation}.
Figure~\ref{fig:motivation}(b) summarizes the mechanism: a biased diffusion prior pulls denoising trajectories toward the frontal mode, and SDS repeatedly converts these biased pseudo-targets into 3D parameter updates.

This imbalance affects SDS through the pseudo-target term $\hat{x}_0^t$ in Eq.~\eqref{eq:sds_pseudotarget}. 
When non-frontal camera views receive pseudo-targets biased toward frontal structure, their gradients no longer provide view-specific supervision. 
Instead, SDS repeatedly pulls different rendered views toward the same high-probability frontal mode, causing trajectory collapse and duplicated front-facing geometry across viewpoints.

This optimization view clarifies why rebalancing camera sampling alone is insufficient: the sampled camera pose may be non-frontal, but the denoising prior can still produce a frontal pseudo-target. 
Rather than fine-tuning the prior, we seek a sampling-time correction at the point where the bias enters SDS: the denoising prediction that forms the pseudo-target. 
Since intermediate diffusion features contain spatially aligned object structure and require no additional predictors, Section~\ref{sec:method} uses them to define a viewpoint-aware structural energy for correcting the denoising trajectory.

\begin{figure*}[htbp]
    \centering
    \includegraphics[width=1.0\linewidth]{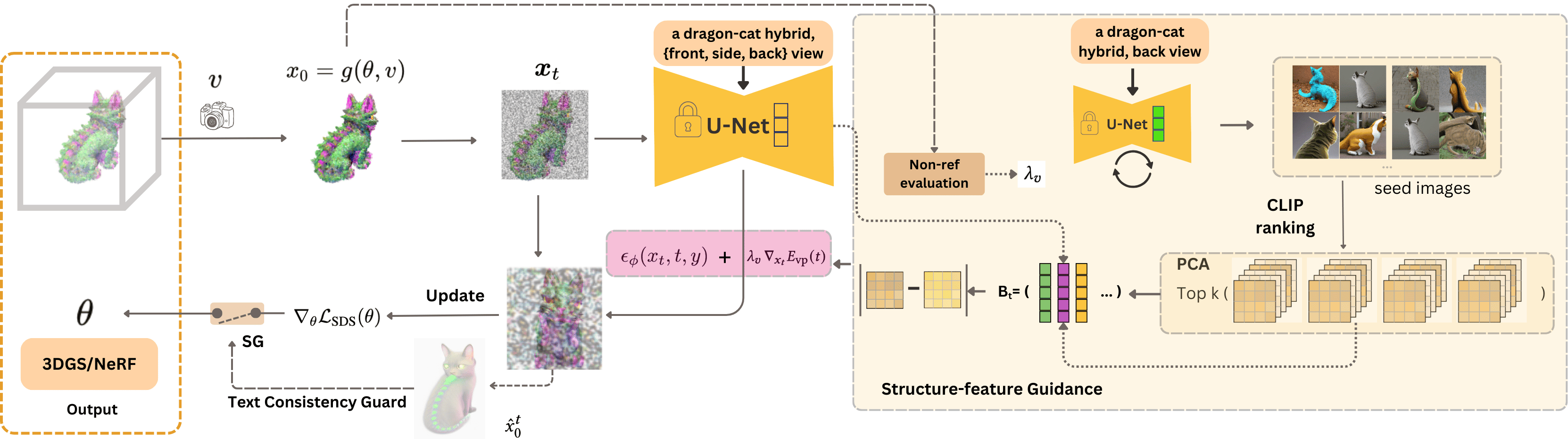} 
    \caption{Overview of our pipeline. We formulate view consistency as \emph{energy-guided sampling}: a PCA-based structural subspace is constructed from intermediate U-Net features to define a viewpoint-aware \emph{structural energy}. Its gradient is injected into the noise prediction at each denoising step, steering the trajectory toward the target viewpoint \emph{without} training or weight updates. An optional text consistency guard can prune misaligned supervision.}
    \label{fig:pipeline}
\end{figure*}

\section{Structural Energy-Guided Sampling}
\label{sec:method}
As outlined in Section~\ref{sec:pre}, we adopt an \emph{energy-guided denoising} perspective, casting view-consistency preservation as minimization of a viewpoint-aware \emph{structural energy} within the iterative denoising process. 
Rather than retraining the backbone diffusion model or rebalancing its data distribution, we introduce a training-free, plug-and-play mechanism that steers each sampling step toward geometry consistent with the target viewpoint.

Self-attention modules in diffusion models contain rich structural representations, capturing spatial layout, object pose, and geometry across different generation stages~\cite{mo2024freecontrol}. 
Beyond their internal role in denoising, these intermediate features can be extracted and repurposed as external guidance signals to influence the sampling trajectory~\cite{epstein2023diffusion}. 
SEGS extracts these features from U\mbox{-}Net decoder self-attention blocks, projects them into a viewpoint-aware PCA subspace, and defines a structural energy that steers denoising toward features from target-view exemplars. 
The following subsections describe the structural subspace (Section~\ref{sec:struct_pca}), the energy definition (Section~\ref{sec:struct_energy}), its denoising update (Section~\ref{sec:eg_update}), and the coupling with SDS/VSD (Section~\ref{sec:coupling}).

\subsection{Structural Subspace via PCA}
\label{sec:struct_pca}
\noindent
This subsection constructs two quantities used by the structural energy: a PCA basis $\mathbf{B}_t$ and target-view structural references $\mathbf{S}_{g,t}$.
Both are derived from intermediate self-attention features of the diffusion U\mbox{-}Net.

\paragraph{Sampling structural features.}
\label{sampling_features}
Given a text prompt $y$ augmented with a target-view token (e.g., \emph{``back view''}), we first generate $N$ auxiliary images using the pretrained diffusion model.
Following~\cite{mo2024freecontrol}, for the $i$-th image at denoising timestep $t$, we extract self-attention \emph{keys} $\mathbf{b}_{t,i}\!\in\!\mathbb{R}^{H\times W\times C}$ from the final layer of the first decoder up\_block of the U\mbox{-}Net (unless otherwise stated).
Such features are known to encode spatially aligned representations~\cite{tumanyan2023plug}.
Collecting across $N$ samples yields a tensor $\mathbf{b}_{t}\!\in\!\mathbb{R}^{N\times H\times W\times C}$.

We then apply Principal Component Analysis (PCA) to $\mathbf{b}_{t}$ (after channel-wise mean centering) to identify dominant directions and form a common feature subspace.
Retaining the first $N_b$ principal components gives
\begin{equation}
\mathbf{B}_t = \text{PCA}(\mathbf{b}_{t}),
\end{equation}
where $\mathbf{B}_t \in \mathbb{R}^{C \times N_b}$.
Unless otherwise noted, $\mathbf{B}_t$ is built at the \emph{current} denoising timestep $t$ to ensure temporal alignment.

Independently, we compute CLIP~\cite{radford2021learning} similarities between each generated image and the target-view text, and select the top-$k$ images.
The corresponding features are aggregated as $\mathbf{F}_t \in \mathbb{R}^{k \times H \times W \times C}$ and projected into the PCA subspace:
\begin{equation}
\mathbf{S}_{g,t} = \mathbf{F}_t \mathbf{B}_t.
\label{eq:structure-guided-feature}
\end{equation}

\begin{figure}[H]
    \centering
    \vspace{-12pt}
    \includegraphics[width=0.82\linewidth]{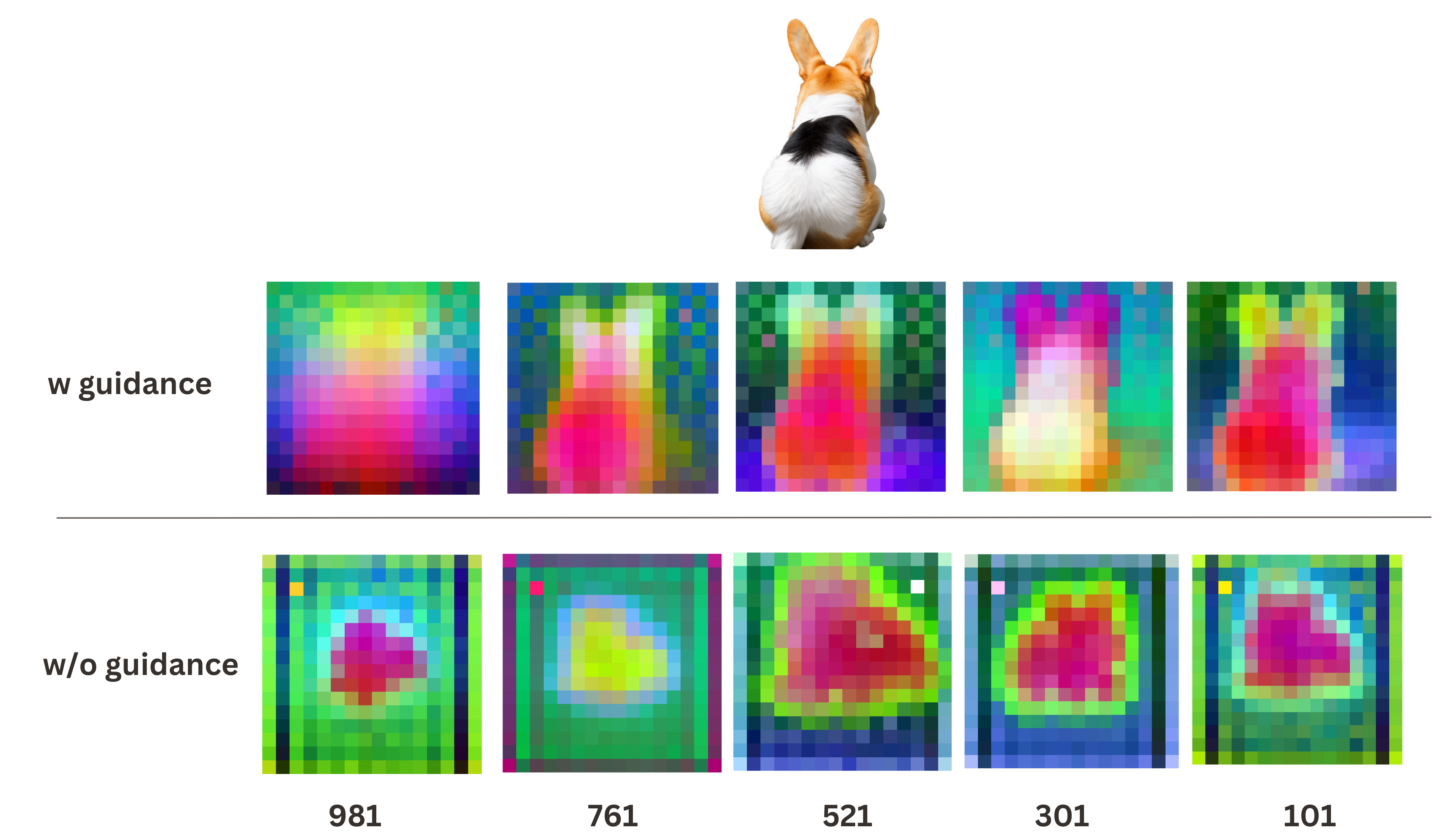}
    \caption{Visualization of intermediate structural features. We apply PCA-based dimensionality reduction to U-Net self-attention features. The rows show the input image, projected features under structural guidance, and projected features without structural supervision, illustrating that the extracted feature maps capture object-level layout and viewpoint-related structure.}
    \label{fig:generation_bias}
\end{figure}

Equation~\eqref{eq:structure-guided-feature} yields $\mathbf{S}_{g,t} \in \mathbb{R}^{k \times H \times W \times N_b}$ as viewpoint-aware structural references distilled from the auxiliary set.
During optimization, we treat $\mathbf{B}_t$ and $\mathbf{S}_{g,t}$ as constants (no gradient flows into the PCA pipeline).
Figure~\ref{fig:generation_bias} visualizes the structural cues captured by the projected intermediate features.

\subsection{Viewpoint Target and Structural Energy}
\label{sec:struct_energy}

\noindent
With the PCA structural basis $\mathbf{B}_t$ and the reference set $\mathbf{S}_{g,t}$ prepared in Section~\ref{sec:struct_pca}, we define a differentiable structural energy $E_{\text{vp}}(t)$ whose gradient will be injected into the denoising update.

\paragraph{Projected current feature.}
At each text-to-3D iteration (Figure~\ref{fig:pipeline}), given the noisy input $x_t$, we extract the U\mbox{-}Net intermediate self-attention feature $\mathbf{f} \in \mathbb{R}^{H \times W \times C}$ and project it onto $\mathbf{B}_t$ (projection along the channel dimension):
\begin{equation}
\mathbf{G}_t \;=\; \mathbf{f}\,\mathbf{B}_t,
\end{equation}
where $\mathbf{G}_t \in \mathbb{R}^{H \times W \times N_b}$ is the subspace-projected representation of the current state.

\paragraph{Structural energy.}
To guide $\mathbf{G}_t$ toward the target-view structural references $\mathbf{S}_{g,t}$ (top-$k$ by CLIP similarity), we define the viewpoint guidance energy $E_{\text{vp}}(t)$ as the average mean squared error between $\mathbf{G}_t$ and each reference $\mathbf{S}_{g,t}[n]$:
\begin{equation}
\begin{aligned}
E_{\text{vp}}(t)
&= \frac{1}{k} \sum_{n=1}^{k}
\frac{1}{H \times W \times N_b}
\sum_{i=1}^{H \times W} \sum_{j=1}^{N_b} \\
&\quad
\left(\mathbf{G}_t[i,j] - \mathbf{S}_{g,t}[n,i,j]\right)^2 .
\end{aligned}
\end{equation}

Here, $i$ and $j$ index the flattened spatial location and PCA-reduced channel, respectively, and $n=1,\dots,k$ indexes the selected references.
Lower $E_{\text{vp}}(t)$ indicates better structural consistency.
Averaging over the top-$k$ references avoids tying the guidance to a single auxiliary sample.
Gradients with respect to $x_t$ are computed via the chain rule:
\[
\nabla_{x_t}E_{\text{vp}}(t)
=\frac{\partial E_{\text{vp}}}{\partial \mathbf f}\frac{\partial \mathbf f}{\partial x_t},
\]
while $\mathbf{B}_t$ and $\mathbf{S}_{g,t}$ remain fixed.

\subsection{Energy-Guided Denoising Update}
\label{sec:eg_update}
\noindent
With $\mathbf{B}_t$, $\mathbf{S}_{g,t}$, and $E_{\text{vp}}(t)$ defined in Section~\ref{sec:struct_energy}, we inject the corresponding gradient into the denoising prediction to steer the sampling trajectory toward the target viewpoint while keeping the diffusion backbone frozen.
At each denoising step $t$, we replace the original noise estimate $\epsilon_{\phi}(x_t,t,y)$ with a viewpoint-corrected version:
\begin{equation}
\hat{\epsilon}_\phi(x_t,t,y)
= \epsilon_{\phi}(x_t,t,y) 
\;+\; \lambda_v(t) \,\nabla_{x_t} E_{\text{vp}}(t),
\label{eq:eg_view}
\end{equation}
where $\lambda_v(t)$ modulates the guidance strength. Since DDPM/DDIM updates subtract $\hat{\epsilon}_\phi$ from the state, adding $+\,\nabla_{x_t}E_{\text{vp}}(t)$ inside $\hat{\epsilon}_\phi$ results in a descent step on $E_{\text{vp}}(t)$ at the state level.

\paragraph{Adaptive guidance.}
In text-to-3D optimization, structural guidance should be stronger while global geometry is being established and relaxed as appearance details stabilize.
We therefore allow the guidance weight to be scheduled adaptively:
\begin{equation}
\lambda_v(t) = \mathcal{F}\bigl(\mathcal{G}(x_0)\bigr),
\end{equation}
where $x_0=g(\theta,v)$ and $\mathcal{G}$ is a no-reference image-quality signal.
The specific BRISQUE-based schedule used in our implementation is described in Section~\ref{sec:implementation-detail}.

\subsection{Coupling with SDS/VSD}
\label{sec:coupling}
\noindent
The proposed guidance operates entirely at sampling time and integrates into standard text-to-3D pipelines as a plug-and-play module.
Substituting the viewpoint-corrected prediction \eqref{eq:eg_view} into the SDS objective (Eq.~\eqref{eq:sds_gradient}) yields
\begin{equation}
\resizebox{0.97\linewidth}{!}{$
\nabla_{\theta} \mathcal{L}_{\text{SDS}}(\theta)
\;\approx\;
\mathbb{E}_{t,\epsilon,v}
\Bigl[
\,\omega(t)\,\bigl(\underbrace{\epsilon_{\phi}(x_t,t,y)
\;+\;\lambda_v(t)\,\nabla_{x_t}E_{\text{vp}}(t)}_{\hat{\epsilon}_\phi(x_t,t,y)}
\;-\;\epsilon_t\bigr)\,\frac{\partial x_0}{\partial\theta}
\Bigr],
$}
\label{eq:combined}
\end{equation}
where an analogous replacement applies to VSD.
In practice, for each rendered view, we compute $E_{\text{vp}}(t)$ and its gradient once per denoising step, form $\hat{\epsilon}_\phi(x_t,t,y)$ as above, and feed it into the outer SDS (or VSD) optimization loop.
This preserves the original backbone and training-free property while imparting explicit viewpoint-aware structural control.

Practical details for the text consistency guard and adaptive guidance scheduling are specified in Section~\ref{sec:implementation-detail}.

\section{Implementation Details}
\label{sec:implementation-detail}
This section summarizes practical components used to instantiate SEGS, including pseudo-supervision filtering and adaptive guidance scheduling.

\subsection{Text Consistency Guard}
We use a lightweight CLIP-based gate to prune pseudo-supervision that is inconsistent with the target-view text.
At each iteration, we encode the pseudo-target $\hat{x}_0^t$ and the view-specific prompt $P_v$ with CLIP and compute the cosine similarity $\sigma$.
During a short warm-up phase, we only log $\sigma$; afterward, we compute an adaptive threshold
\[
\tau=\alpha\,\sigma_{\min}+(1-\alpha)\,\sigma_{\text{mean}},\qquad \alpha=0.5,
\]
and discard supervision with $\sigma<\tau$.
This reduces misaligned gradients while retaining most valid signals.

\subsection{Adaptive Guidance Scheduling}
\paragraph{BRISQUE schedule.}
To allocate supervision when shape formation matters most, we modulate the structural guidance weight by a no-reference quality score.
Let $b_t$ be the BRISQUE score (higher is worse) of the current renderings at iteration $t$.
We smooth $b_t$ using an exponential moving average (EMA):
\[
\tilde b_t=\beta\,\tilde b_{t-1}+(1-\beta)\,b_t,\quad \beta=0.9,
\]
normalize it into $q_t\in[0,1]$ with a running range (e.g., percentile-based to avoid outliers):
\[
q_t=\mathrm{clamp}\!\left(\frac{\tilde b_t-b_{\min}}{b_{\max}-b_{\min}},\,0,\,1\right),
\]
and set the structural guidance weight as:
\[
\lambda_v(t)=\lambda_{\min}+(\lambda_{\max}-\lambda_{\min})\,q_t.
\]

\begin{figure}[H]
  \centering
  \includegraphics[width=\linewidth]{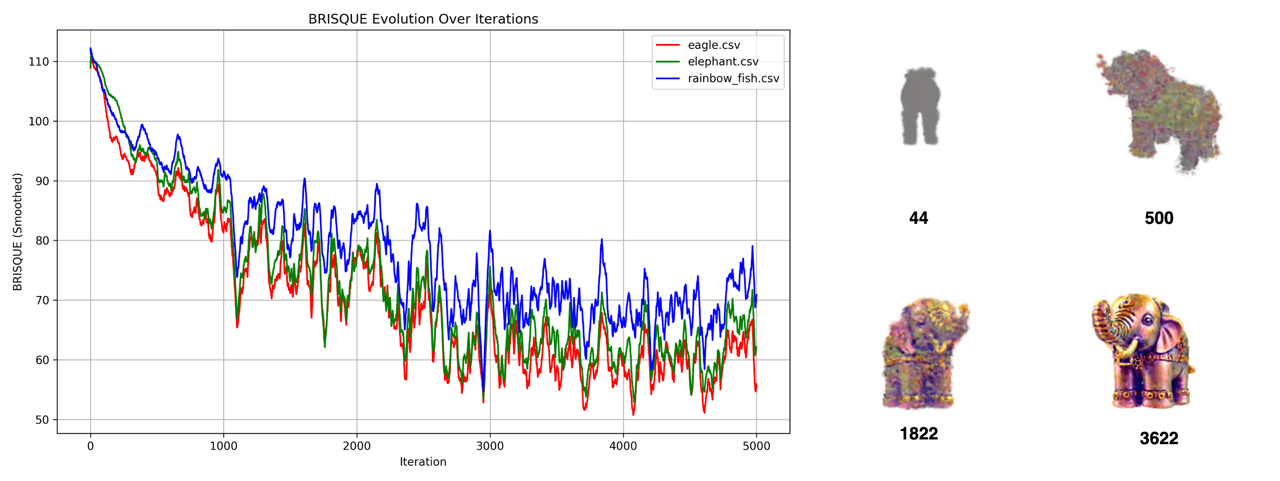}
  \caption{\textbf{BRISQUE-guided scheduling.}
  Left: smoothed BRISQUE over iterations for three prompts shows a consistent decline.
  Right: representative renderings with their BRISQUE scores at different iterations.
  This trend motivates applying stronger structural guidance early and relaxing it as renderings improve.}
  \label{fig:brisque}
\end{figure}

Hence, early low-quality renderings (large BRISQUE) receive stronger guidance, and the weight decays as quality improves.
Figure~\ref{fig:brisque} shows that BRISQUE curves for different prompts follow a similar downward trend, supporting this schedule.

\begin{figure}[H]
  \centering
  \includegraphics[width=\linewidth]{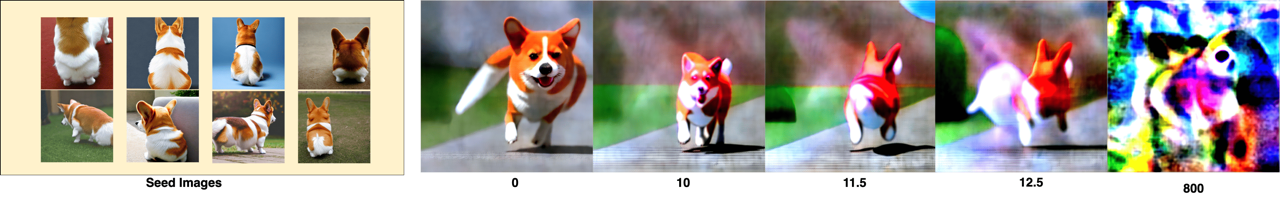}
  \caption{\textbf{Effect of $\lambda_v$ (``a corgi is running'').}
  Left: seed images used to build structural references.
  Middle: results with increasing $\lambda_v$ (0, 10, 11.5, 12.5).
  Right: an extreme setting ($\lambda_v{=}800$) leads to collapse.
  Moderate guidance helps correct viewpoint/structure; overly large guidance overrides the diffusion prior.}
  \label{fig:lambda}
\end{figure}

\paragraph{Guidance-weight range.}
We analyze the effect of the guidance strength using the prompt \emph{``a corgi is running''}.
Figure~\ref{fig:lambda} varies $\lambda_v$ while keeping other settings fixed.
When $\lambda_v=0$, no structural correction is applied, and the generation tends to stay near the prior’s frontal mode.
As $\lambda_v$ increases, around $\lambda_v\!\approx\!11.5$, the method begins to noticeably steer the trajectory toward the desired structure/view.
However, extremely large values (e.g., $\lambda_v=800$) overwhelm the diffusion prior, leading to collapse.
In practice, we therefore (i) use the BRISQUE schedule above to adjust $\lambda_v(t)$ over time, and (ii) cap the maximum value $\lambda_{\max}$ conservatively below the collapse regime (in the corgi example, below the hundreds range).

\begin{figure*}[htbp]
    \centering
    \includegraphics[width=1.0\linewidth]{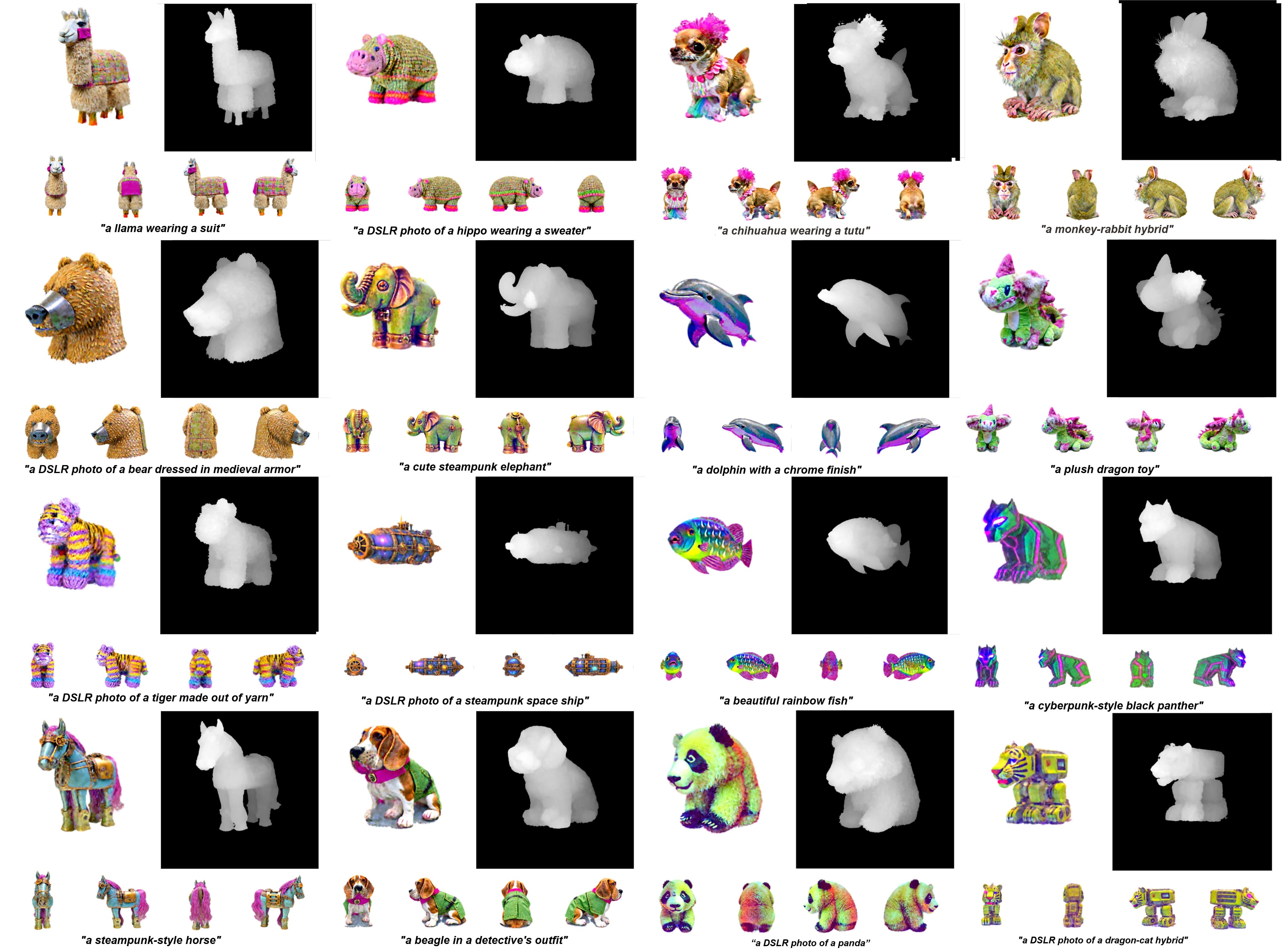}
    \caption{\textbf{Examples generated by SEGS.} The results show diverse prompts with consistent geometry and detailed textures.}
    \label{fig:main}
\end{figure*}

\begin{figure}[t]
    \centering
    \includegraphics[width=0.9\linewidth]{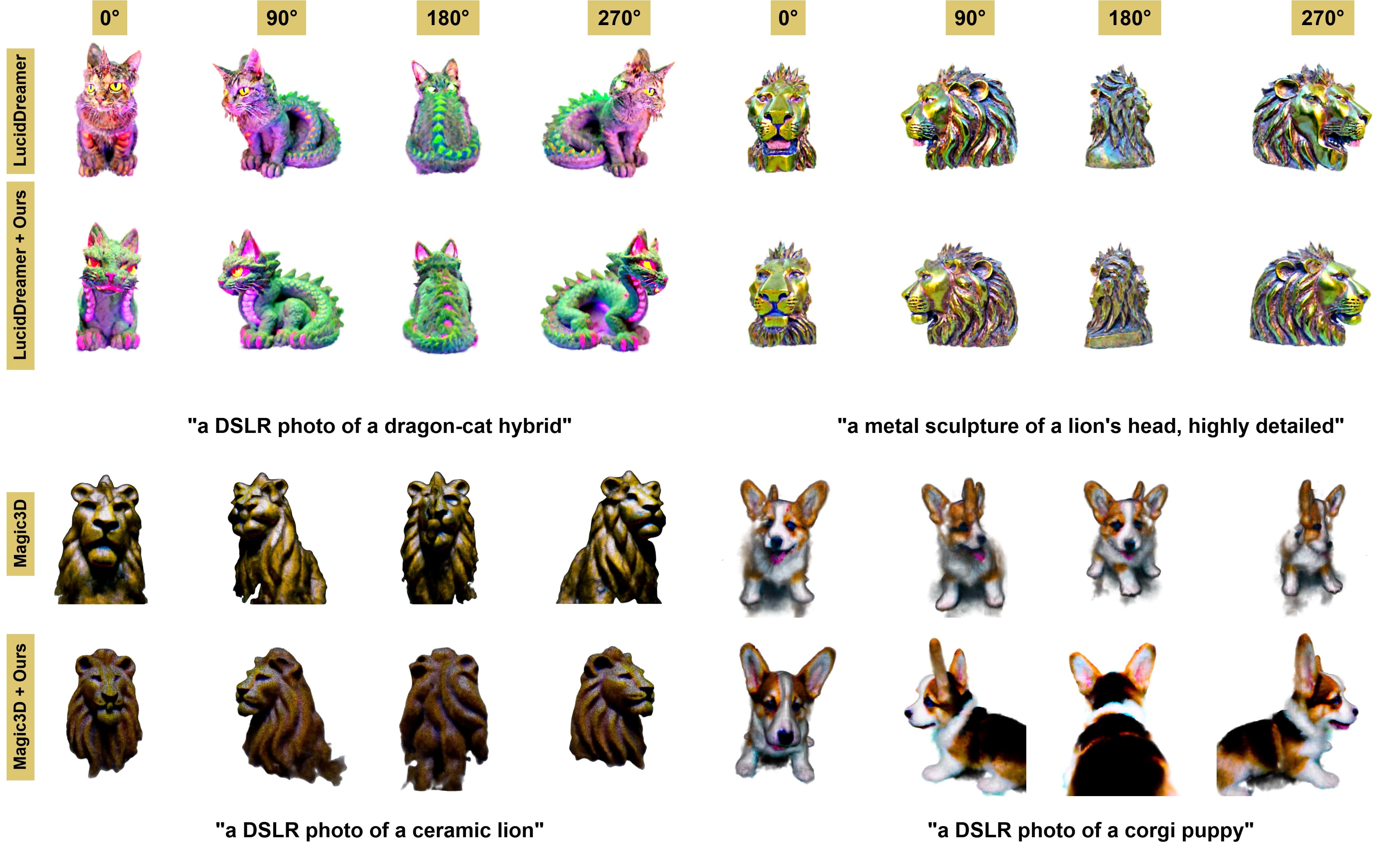}
    \caption{\textbf{Comparison with baseline methods in text-to-3D generation.} Adding SEGS to each baseline improves geometric consistency and reduces Janus artifacts under the same generation setting.}
    \label{fig:example}
\end{figure}

\section{Experiments}
\label{sec:exp}
We evaluate the proposed SEGS on the DreamFusion prompt library~\cite{poole2022dreamfusion}.
Both qualitative and quantitative comparisons are provided, showing improved \emph{view consistency}.
We also conduct ablations to assess the contribution of each component within SEGS.

\subsection{Experimental Setup}
\label{sec:exp_implementation}
\noindent \textbf{Prompt handling and view bins.}  
To obtain object-centric supervision, we simplify complex prompts using a large language model. For each prompt from the DreamFusion library, the main object token is extracted and used to generate auxiliary features and initialize 3D representations.
To specify underrepresented viewpoints, we partition azimuth into three coarse bins: front ($[-60^\circ,60^\circ)$), side ($[-120^\circ,-60^\circ)\cup[60^\circ,120^\circ)$), and back ($[-180^\circ,-120^\circ)\cup[120^\circ,180^\circ)$), and focus guidance on the \emph{back} bin.
View-specific auxiliary and evaluation prompts are formed by combining the simplified object token with viewpoint descriptors such as ``back view.''

\noindent \textbf{SEGS settings.}
Unless otherwise stated, we instantiate SEGS as described in Secs.~\ref{sec:method} and~\ref{sec:implementation-detail}.
We extract self-attention \emph{keys} from the final layer of the first decoder up\_block of the U\mbox{-}Net and apply channel-wise mean centering before PCA.
For each prompt, we generate 20 auxiliary ``back view'' images with Stable Diffusion v1.4, set the PCA dimension to $N_b{=}64$, and select top-$k$ samples with $k{=}3$.
The text consistency guard and BRISQUE-based guidance schedule follow Section~\ref{sec:implementation-detail}.
All experiments are run on a single NVIDIA A100.

\noindent \textbf{Janus Rate (JR).}  
To measure geometric consistency in multi-view 3D generation, we define the Janus Rate (JR) as the percentage of generated objects whose rendered views exhibit inconsistent or duplicated front-facing features, such as multiple faces or misplaced facial parts. We manually inspect renderings from multiple azimuth angles and mark an object as a failure if Janus artifacts appear in any non-frontal view.

\noindent \textbf{View-Dependent CLIP Score (View-CS).}  
The conventional CLIP Score measures the semantic similarity between a generated image and a full-text prompt. However, it cannot accurately reflect viewpoint inconsistencies, since front-view features tend to dominate similarity regardless of viewpoint correctness.

To address this, we propose the View-Dependent CLIP Score (View-CS), which isolates viewpoint fidelity from object semantics. Instead of using the full prompt, we compute CLIP similarity between rendered images and standalone viewpoint descriptors: \textit{``front view,'' ``side view,''} and \textit{``back view.''} For each 3D object, we render images from four azimuth angles (\(0^\circ\), \(90^\circ\), \(180^\circ\), and \(270^\circ\)) and match them with \textit{``front view,'' ``side view,'' ``back view,''} and \textit{``side view,''} respectively. The final View-CS is obtained by averaging these four similarities, providing an objective measure of how well the generated geometry aligns with the intended camera direction.

\subsection{Text-to-3D Generation}
\noindent \textbf{Qualitative Comparison.}
We primarily compare SEGS with representative text-to-3D baselines, including a vanilla SDS implementation via ThreeStudio~\cite{liu2023threestudio} and LucidDreamer~\cite{liang2024luciddreamer}. 
All methods are distilled from Stable Diffusion v1.4 on the same compute. 
We focus on SDS-style pipelines since SEGS is designed as a \emph{training-free, plug-and-play} module that directly enhances such frameworks; fine-tuned paradigms like Zero-1-to-3 or MVDream rely on additional curated data and retraining, which falls outside our scope. 
Figure~\ref{fig:main} shows representative SEGS generations across diverse prompts.
As shown in Figure~\ref{fig:example}, SEGS visibly reduces Janus artifacts and improves view consistency relative to the baselines.

\subsection{Quantitative Comparison}
\begin{table}[t]
\centering
\begin{tabular}{lcc}
\toprule
\textbf{Method} & \textbf{JR (\%)$\downarrow$} & \textbf{View-CS $\uparrow$} \\
\midrule
LucidDreamer~\cite{liang2024luciddreamer} & 58.06 & 30.16 \\
LucidDreamer+SEGS (Ours) & \textbf{48.39} & \textbf{30.95} \\
\midrule
Magic3D~\cite{lin2023magic3d} & 68.18 & 32.17 \\
Magic3D+SEGS (Ours) & \textbf{56.36} & \textbf{32.85} \\
\midrule
DreamFusion~\cite{poole2022dreamfusion} & 72.73 & 32.85 \\
DreamFusion+SEGS (Ours) & \textbf{63.64} & \textbf{33.29} \\
\bottomrule
\end{tabular}
\vspace{6pt}
\caption{Comparison of geometric consistency between baselines and SEGS-enhanced variants. SEGS lowers JR and improves View-CS across the evaluated baselines, indicating improved viewpoint alignment.}
\label{table1}
\end{table}

Previous research in 3D generation has lacked standardized evaluation metrics for viewpoint-dependent inconsistencies, particularly those characteristic of the Janus problem. 
Using JR and View-CS as defined in Section~\ref{sec:exp_implementation}, we evaluate 124 prompts and render each generated asset from four uniformly distributed azimuth angles.
As reported in Table~\ref{table1}, SEGS reduces JR and improves View-CS across all three baseline methods, suggesting enhanced viewpoint fidelity without compromising geometric consistency.

\subsection{Ablation Study}
To verify the effectiveness of each component in SEGS, we carry out ablation studies.

\begin{figure}[H]
    \centering
    \includegraphics[width=1\linewidth]{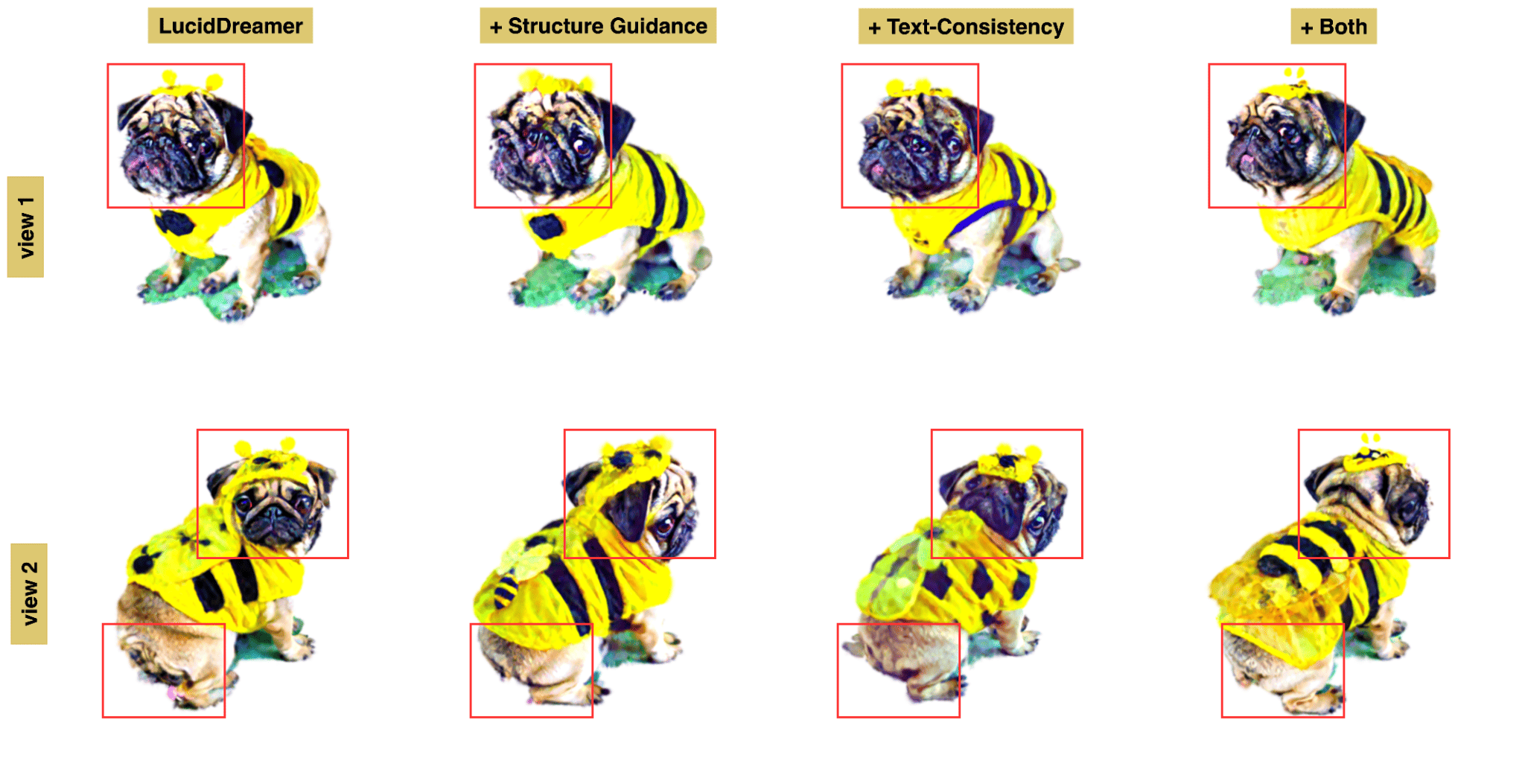}
    \vspace{-18pt}
    \caption{Ablation study on the guidance modules using LucidDreamer as the baseline. Under a back-view rendering, \textit{SEGS} reduces front-facing features, while the \textit{Text Consistency Guard} alone provides limited suppression. Combining \textit{SEGS + Text Consistency Guard} further improves alignment with the intended back-view perspective.}
    \label{fig:effect_guidance}
\end{figure}

\noindent \textbf{Effect of Guidance.} 
As shown in Figure~\ref{fig:effect_guidance}, we visualize renders at a fixed back-view camera to compare guidance variants. In both the baseline and the \emph{Text Consistency Guard}-only settings, front-facing facial cues still leak into back views, revealing a strong front-view bias. Applying \emph{Structural Energy Guidance} suppresses these artifacts but may leave mild geometric inconsistencies around the head region. In contrast, combining structural guidance with the text consistency guard removes the residual leaks and yields more consistent, viewpoint-accurate 3D representations.

\begin{figure}[H]
    \centering
    \includegraphics[width=0.68\linewidth]{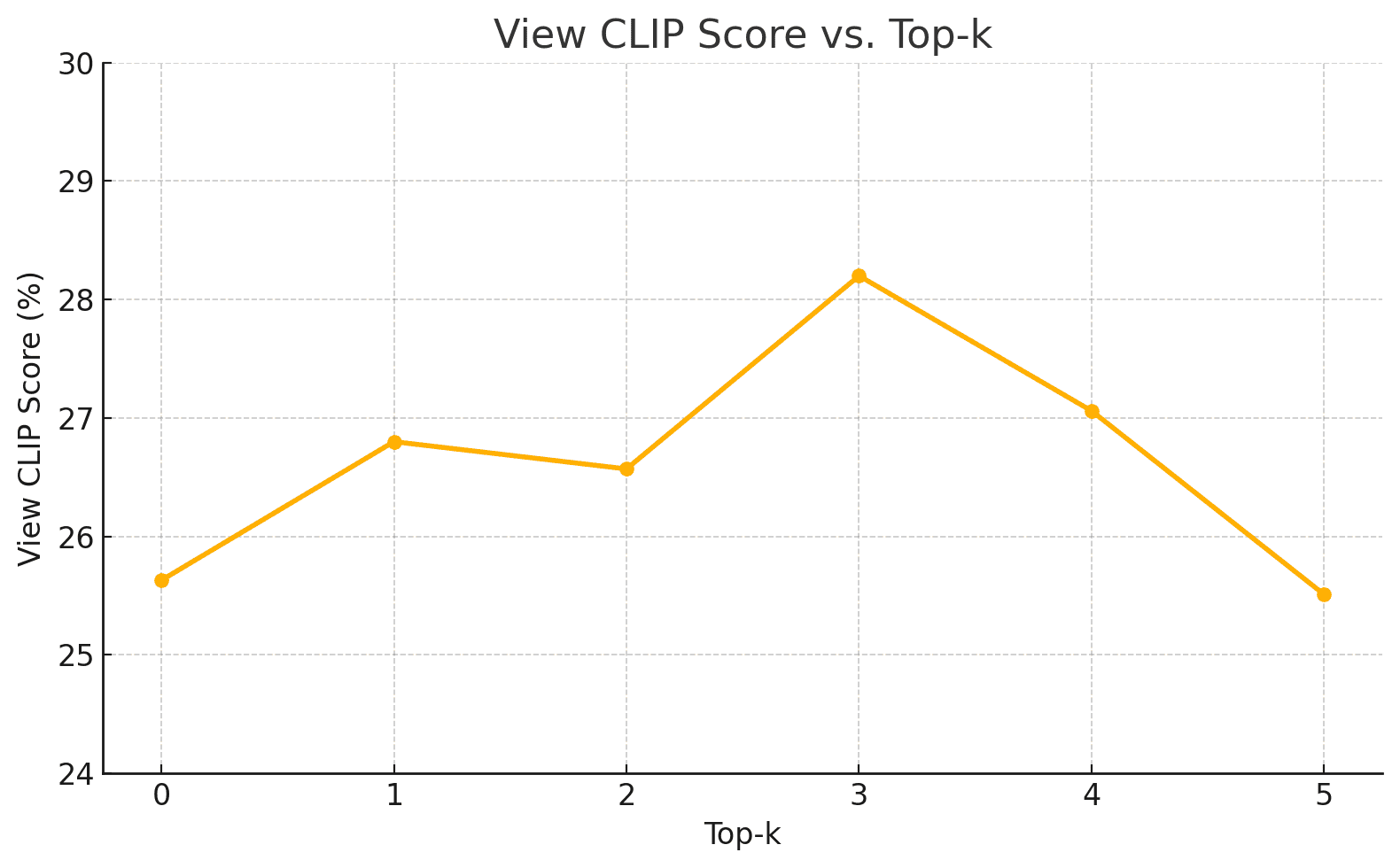}
    \caption{Effect of top-$k$ on View-CS. Increasing top-$k$ improves consistency up to top-$3$, after which performance decreases.}
    \label{fig:top-k}
\end{figure}

\noindent \textbf{Top-$k$.}
Top-$k$ refers to the number of CLIP-selected auxiliary images used as target-view structural references after constructing the shared PCA basis. We evaluate its impact using View-CS across the front, side, and back rendered views of the generated 3D objects. As shown in Figure~\ref{fig:top-k}, View-CS initially rises as top-$k$ increases and then drops. The highest score is observed at top-$3$, indicating it is the best setting in this ablation.

\noindent \textbf{Comparison Across Diffusion Versions.}  
To examine how backbone models influence the Janus problem, we test both Stable Diffusion v1.4 and v2.1 using DreamFusion. The prompt ``a DSLR photo of a corgi puppy'' is randomly sampled five times. Table~\ref{tab:sd_versions} reports the average Janus Rate under each configuration.
While SD2.1 slightly reduces the frequency of Janus artifacts, both versions show lower JR when SEGS is applied. This suggests that the training-free guidance remains beneficial across these two backbone versions.

\begin{table}[htbp]
\centering
\begin{tabular}{lcc}
\toprule
\textbf{Method} & \textbf{SD1.4} & \textbf{SD2.1} \\
\midrule
DreamFusion & 100\% & 60\% \\
DreamFusion + SEGS (Ours) & \textbf{40\%} & \textbf{20\%} \\
\bottomrule
\end{tabular}
\vspace{6pt}
\caption{Janus Rate (JR↓) on DreamFusion with and without SEGS for SD-1.4 and SD-2.1 using one prompt and five seeds. SEGS reduces JR for both backbone versions in this setting.}
\label{tab:sd_versions}
\end{table}

\begin{figure}[H]
    \centering
    \includegraphics[width=0.82\linewidth]{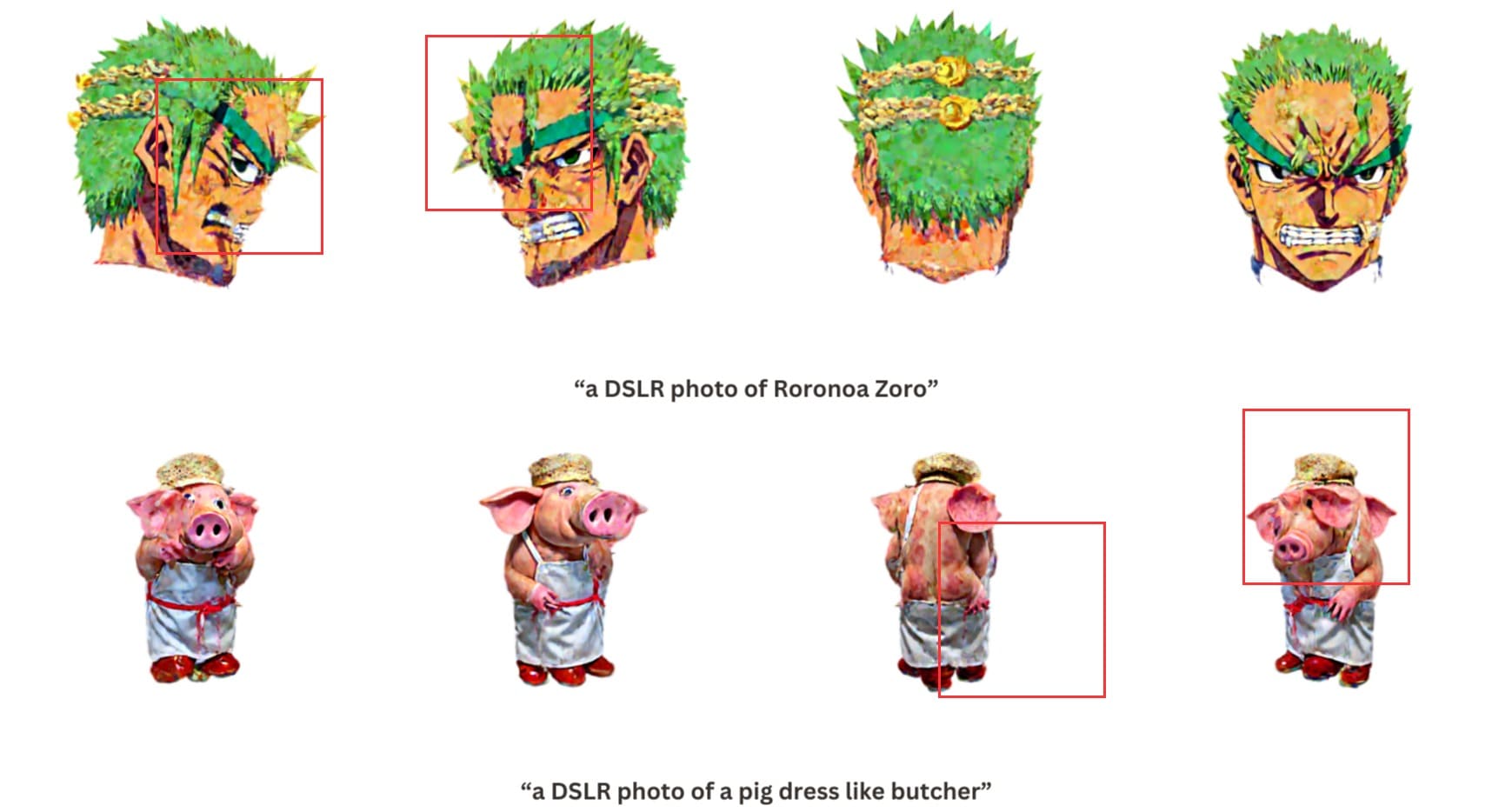}
    \caption{Failure cases at front--side transitions. \textbf{Red boxes} highlight residual frontal cues (e.g., eyes, ears, snout) that persist around lateral contours. These errors likely stem from ambiguous side-view evidence in the 2D prior and \emph{weaker structural targets at intermediate azimuths}, as our reference set focuses on the back.}
    \label{fig:failure_case2} 
\end{figure}

\noindent \textbf{Failure Case Analysis.}
While SEGS reduces Janus artifacts, failures still occur near front--side transition angles (Figure~\ref{fig:failure_case2}), where faint frontal cues may remain along lateral contours.
We attribute this to (i) ambiguity in side-view evidence in the 2D prior and (ii) weaker structural targets at intermediate azimuths because the current reference set concentrates on the back bin.

\section{Conclusion}
\label{sec:conclusion}
In this paper, we identify viewpoint bias in the training images of diffusion models as a key factor underlying the Janus problem in text-to-3D generation. 
To address this issue, we propose SEGS, a training-free framework that combines viewpoint-aware structural energy guidance with a lightweight Text Consistency Guard. 
SEGS reduces Janus artifacts while preserving appearance fidelity, and integrates seamlessly as a plug-and-play module into existing text-to-3D pipelines. 
Looking ahead, we observe that finer-grained cues, such as facial details, tend to emerge in later U-Net layers; exploring guidance that explicitly steers trajectories away from these late-stage frontal features may offer a promising direction for further mitigating Janus artifacts.

\section*{Declarations}
\noindent \textbf{Conflict of interest.} 
The authors declare they have no conflict of interest.

\section*{Data Availability}
No new datasets were generated in this study. The public data sources used for prompts and viewpoint-bias analysis are described in the paper. Derived evaluation records are available from the corresponding author upon reasonable request.
The source code for SEGS is publicly available at \url{https://github.com/QZhang2111/SEGS}.

\newpage
\bibliography{sn-bibliography} 
\end{document}